\documentclass[final]{cvpr}
\usepackage{comment}
\usepackage{times}
\usepackage{epsfig}
\usepackage{graphicx}
\usepackage{amsmath}
\usepackage{amssymb}
\usepackage{url}
\usepackage[ruled,vlined]{algorithm2e}
\usepackage[pagebackref=true,breaklinks=true,colorlinks,bookmarks=false]{hyperref}

\def\cvprPaperID{6230} % *** Enter the CVPR Paper ID here
\def\confYear{CVPR 2021}

\usepackage{xcolor}
\usepackage{multirow}

\begin{document}

%%%%%%%%% TITLE
\title{Automatic Vertebra Localization and Identification in CT by Spine Rectification and Anatomically-constrained Optimization}
\author{
 % Authors
 \thanks{This work was done when Fakai Wang interned at PAII Inc.}
 Fakai Wang\textsuperscript{\rm 1,\rm 2},
 Kang Zheng\textsuperscript{\rm 1},
 Le Lu\textsuperscript{\rm 1},
 Jing Xiao\textsuperscript{\rm 3 },
 Min Wu\textsuperscript{\rm 2 },
 Shun Miao\textsuperscript{\rm 1 } \\
% {
 % Affiliations
 \textsuperscript{\rm 1} PAII Inc., Bethesda, Maryland, USA  \\ 
 \textsuperscript{\rm 2} University of Maryland College Park, USA \\
\textsuperscript{\rm 3} Ping An Technology, Shenzhen, China}

\maketitle
%%%%%%%%% ABSTRACT
\begin{abstract}
	Accurate vertebra localization and identification are required in many clinical applications of spine disorder diagnosis and surgery planning. However, significant challenges are posed in this task by highly varying pathologies (such as vertebral compression fracture, scoliosis, and vertebral fixation) and imaging conditions (such as limited field of view and metal streak artifacts). This paper proposes a robust and accurate method that effectively exploits the anatomical knowledge of the spine to facilitate vertebra localization and identification. A key point localization model is trained to produce activation maps of vertebra centers. They are then re-sampled along the spine centerline to produce spine-rectified activation maps, which are further aggregated into 1-D activation signals. Following this, an anatomically-constrained optimization module is introduced to jointly search for the optimal vertebra centers under a soft constraint that regulates the distance between vertebrae and a hard constraint on the consecutive vertebra indices. When being evaluated on a major public benchmark of 302 highly pathological CT images, the proposed method reports the state of the art identification (id.) rate of 97.4\%, and outperforms the best competing method of 94.7\% id. rate by reducing the relative id. error rate by half.
\end{abstract}

%%%%%%%%% BODY TEXT
\section{Introduction}

\label{sec:intro}

Localization and identification of spine vertebrae in 3-D medical images are key enabling components for computer-aided diagnosis of spine disorders~\cite{pickhardt2020automated}. As a prerequisite step of downstream applications, high accuracies of vertebra localization and identification are frequently demanded. In recent years, many studies have been reported to address this problem, with substantial progress on public benchmarks (e.g., the SpineWeb~\cite{csi2014}. However, due to the similar appearances of the spine vertebrae, it remains a daunting task to identify vertebrae with a very high accuracy that meets the requirements of clinical applications. 

\begin{figure}[]
	\begin{minipage}[b]{1.0\linewidth}
		\centering
		\centerline{\includegraphics[width=8.5cm]{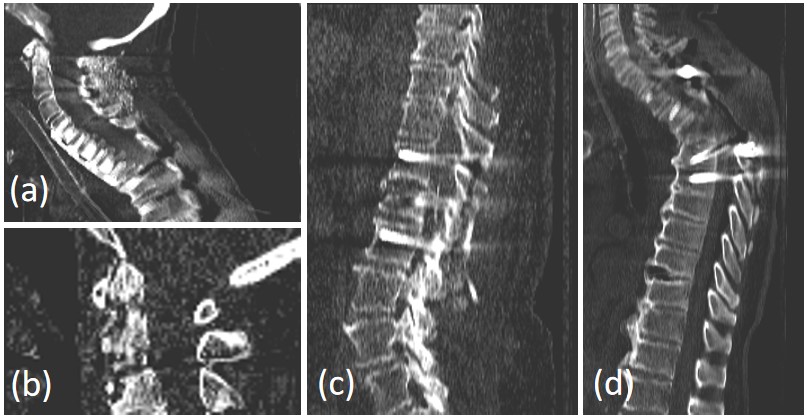}}
	\end{minipage}
	\caption{Example spine CT images from the SpineWeb benchmark dataset demonstrating the challenges. (a) small field of fiew, (b) low image quality, (c,d) metal implants and severe compression fracture.}
	\label{fig:representative_challenges}
\end{figure}

The challenges in distinguishing vertebrae with similar shapes/appearances are well recognized by the research community~\cite{liaohaofu2018,yangdong2017,yizhichen}. Multiple methods have been proposed to address them by exploiting the anatomical prior knowledge: 1) the spatial order of vertebrae, and 2) the distance between neighboring vertebrae. Spine anatomical knowledge is incorporated into neural networks implicitly using Bi-RNN~\cite{liaohaofu2018}, or explicitly using an information aggregation layer considering the spatial distribution prior of the vertebrae~\cite{yangdong2017}. The anatomical prior has also been used to post-process the neural network output~\cite{yizhichen}. While steady performance improvements are observed in these works, the anatomical knowledge is still not fully utilized. In particular, anatomy-inspired network architectures like Bi-RNN~\cite{liaohaofu2018} rely on the network to learn the anatomical prior without the guaranteed respect to the prior. Building the anatomical knowledge into a network layer~\cite{yangdong2017} or the optimization target~\cite{yizhichen} makes a compromise that turns the hard constraint (which should be strictly enforced, e.g., the spatial order) into soft constraints that can be violated. As a result, previous methods may produce physically implausible predictions (e.g., vertebrae in reversed order, multiple occurrences of the same vertebra). 

Furthermore, while previous methods employ the information exchange mechanisms (e.g., Bi-RNN~\cite{liaohaofu2018} and message passing~\cite{yangdong2017}) to bring the global context into vertebra label classification scores, the vertebra label is still classified individually at the output stage for each vertebra without imposing the anatomical constraints. Therefore, these methods completely depend on the information exchange mechanisms to capture and regulate the spatial relationships between vertebrae. Existing fusion mechanisms include 1) recurrent neural network~\cite{liaohaofu2018}, which \textit{encourages} the message passing between vertebrae in a softly learned way instead of \textit{enforcing} it in an anatomy coherent manner; 2) aggregation of the neighboring vertebrae's activation maps~\cite{yangdong2017} following the vertebra distance prior, which is only reliable for short-range relationships, leaving the global anatomical knowledge insufficiently exploited. A specific optimization formulation is used in~\cite{yizhichen} to jointly label the vertebrae by formulating a global objective function. However, the Markov modeling of vertebra labels employed in~\cite{yizhichen} is still limited to capture the short-range relationships and the error accumulates with the Markov steps. 

In this work, we propose a vertebra localization and identification method that jointly labels all vertebrae with anatomical constraints to effectively utilize the anatomical knowledge and achieve the optimal robustness. In particular, a key point localization U-Net~\cite{ronneberger2015u} is trained to predict activation maps for the 26 vertebra centers. Along the automatically calculated spine centerline, the activation maps are warped to rectify the spine and aggregated to form novel 1-D vertebra activation signals. Vertebra localization and identification tasks are then formulated as an optimization problem on the 1-D signals. The spatial order of the vertebrae is guaranteed using a hard constraint to limit the optimization search space. The prior knowledge of the distance between vertebrae is integrated via a soft constraint, i.e., a regularization term in the objective function. The labels for all vertebrae are searched jointly in the constrained search space, which allows global message passing among the vertebrae and ensures the anatomical plausibility of the results. We evaluate our method on a main public benchmark from SpineWeb with a training set of 242 CTs and a testing test of 60 CTs. Our method reports the new state-of-the-art identification rate of 97.4\%, significantly outperforming the previously best competing method~\cite{yizhichen} that achieves a rate of 94.7\%.

In summary, our contributions are four-fold. {\bf 1)} We propose a simple yet effective approach to aggregate 3-D vertebra activation maps into 1-D signals so that the complexity of the task is significantly reduced. {\bf 2)} We exploit the spatial order of the vertebrae as a hard constraint of the optimization search space, which anatomically ensures plausible outputs.  {\bf 3)} We introduce the vertebra distance prior as a soft constraint in optimization of the objective function, flexibly leveraging the relation between vertebrae.  {\bf 4)} Our method achieves the new state-of-the-art performance by improving the identification accuracy from 94.7\% to 97.4\% and equivalently cutting the error rate by half.

\section{Related Work}
\label{sec:related}
Early works on vertebra localization and identification before the era of deep learning rely on hand-crafted low-level image features and/or a priori knowledge. Glocker \etal~\cite{glockerRegressionForest2012} propose to use regression forests and probabilistic graphical models to handle arbitrary field-of-view CT scans. They ~\cite{Glocker2013sparseAnnotations} further transform the sparse centroid annotations into dense probabilistic labels for classifier training. Zhan \etal~\cite{zhan2012robust} use a hierarchical strategy to learn detectors dedicated to distinctive vertebrae and non-distinctive vertebrae. While these methods produce promising results, due to the limited modeling power of hand-crafted features, they lack robustness and produce erroneous results on challenging pathological images. In addition, they fail to exploit the global contextual information to facilitate vertebra identification. 

Deep neural networks are employed to detect spine vertebrae and achieve substantially improved performance. A few publications~\cite{chenhao,suzani2015fast} employ convolutional neural network (CNN) to directly detect the vertebra centers. Fully convolutional network (FCN)~\cite{Trevor_FCN} has also been adopted for the vertebra center detection task~\cite{liaohaofu2018,yangdong2017,Qin2020ResidualBM}. These methods achieve the vertebra localization and identification tasks jointly in one stage. Others employ multiple stages to locate and identify the vertebrae, which can be categorized into top-down~\cite{jakubicek19learningbased,McCouat2019VertebraeDA} or bottom-up strategies. A top-down scheme locates the whole spine first and detects individual vertebrae next. A bottom-up strategy first detects the landmarks of all vertebrae and then classify them into the respective vertebrae~\cite{windsor2020convolutional,sonsbee19ISBIweaklylabeled}. 

\begin{figure*}[bt!]
	\centering
	\includegraphics[width=0.95\linewidth]{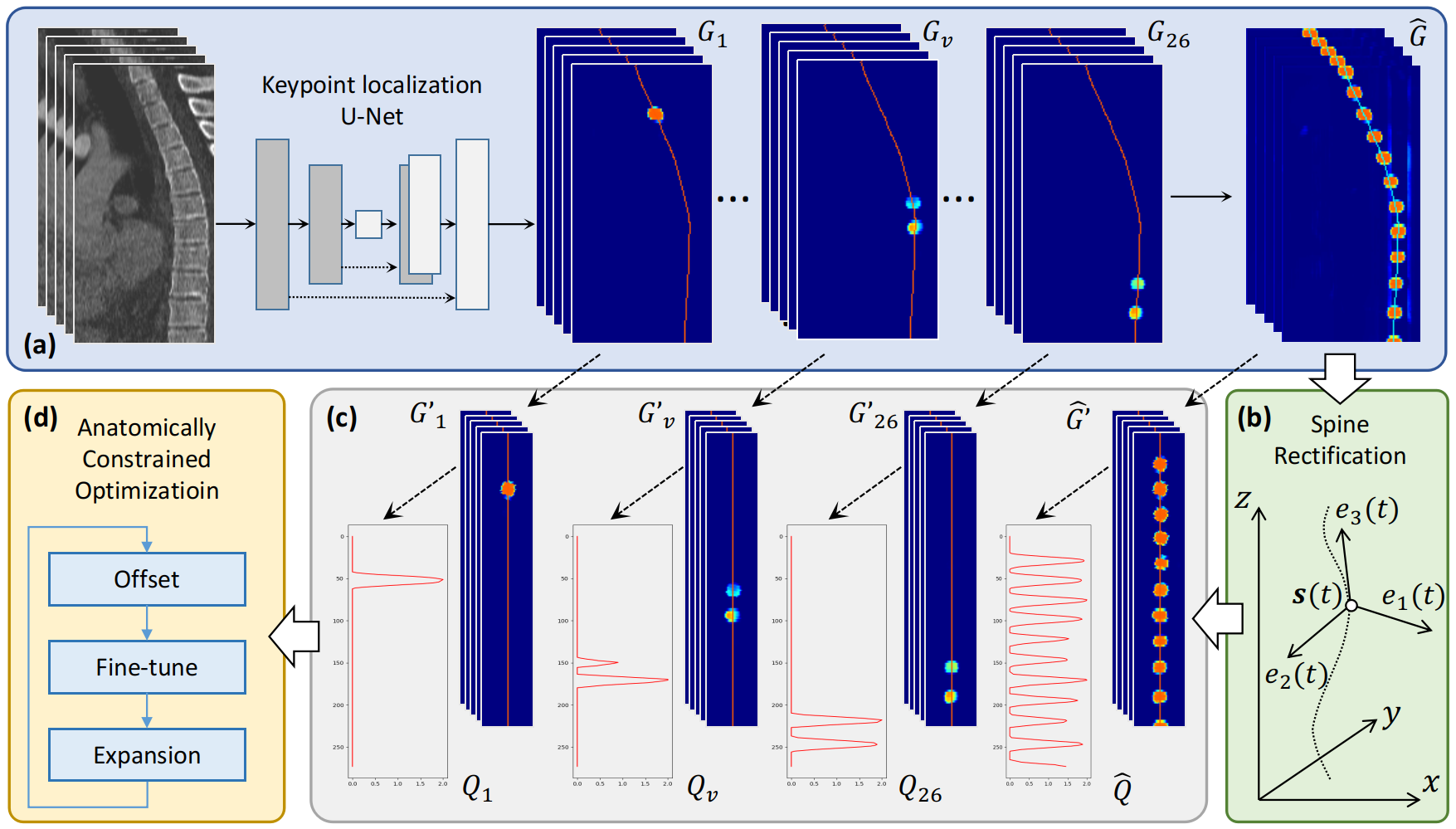}
	\caption{Overview of the proposed system. (a) 26 vertebra activation maps $\{G_v\}$ produce by the key point localization nnU-Net, and the all vertebrae activation map $\hat{G}$ produced by aggregating $\{G_v\}$. The centerline of the spine is marked in the activation maps. (b) Spine rectification operator derived from the spine centerline $\mathbf{s}(t)$ in $\hat{G}$. (b) Spine rectified activation maps $G_v'$ and $\hat{G}$ produced by applying spine rectification on $\{G_v\}$ and $\hat{G}$. (c) The 1-D vertebra activation signals $\{Q_v\}$ and $\hat{Q}$ produced by spatially aggregating $\{G_v\}$ and $\hat{G}$. (d) The anatomically constrained optimization module applied on the 1-D activation signals.}
	\label{fig:system}
\end{figure*}

Many techniques have studied the use of a priori knowledge of spine anatomy to facilitate vertebra localization and identification~\cite{chenhao,zhan2012robust,liaohaofu2018,yangdong2017,yizhichen,Qin2020ResidualBM,btrfly2018}. Domain expert knowledge is used to categorize vertebrae into anchor and bundle sets which are treat differently in the process~\cite{zhan2012robust}. Markov modeling is adopted to label vertebrae by preserving the consecutive order~\cite{yizhichen}. Various attempts have been made to automatically learn the knowledge in a data-driven manner~\cite{liaohaofu2018,yangdong2017,Qin2020ResidualBM,btrfly2018}. Bi-directional recurrent neural network (RNN) is adopted to enable the model to capture the spatial relations of predictions in different regions~\cite{liaohaofu2018,Qin2020ResidualBM}. A message passing mechanism is used to exploit the prior distribution of the distance between vertebrae to regulate the prediction~\cite{yangdong2017}. Adversarial learning has also been employed to encode and impose the anatomical prior~\cite{btrfly2018}. The multi-stage methods ~\cite{jakubicek19learningbased,McCouat2019VertebraeDA,windsor2020convolutional,sonsbee19ISBIweaklylabeled} embed the knowledge of the spine anatomy in their top-down and bottom-up representations.

\section{Methods}
\label{sec:Methodology}

Given a CT image/scan of size ${W\times H\times L}$, denoted as $I \in \mathbb{R}^{W\times H\times L}$, the goal of vertebra localization and identification is to detect the centers of the spine vertebrae that are present in $I$ and identify their labels. There are in total 26 vertebra labels, including 7 cervical, 12 thoracic,  5 lumbar, and 2 sacrum vertebrae. The model takes the image $I$ as input and outputs the centers of the detected vertebrae $\mathbf{P} = \{x_v, y_v, z_v \}, v \in V$, where $V \subseteq \{1, 2, \dots 26\}$ denotes the indices of any detected vertebrae. For all images in training, the vertebra center annotations $\mathbf{P}$ are provided. Our proposed system consists of three steps: 1) training a U-Net key point detection model to estimate 26 vertebra activation maps; 2) spine rectification to produce 1-D activation signal; 3) anatomically constrained optimization to detect vertebra centers from the 1-D signal.

\subsection{Generation of Vertebra Activation Map}
\label{ssec:probability}

In the first step, we train a key point localization model using U-Net as the backbone network to produce activation maps of 26 vertebra centers. This model is trained using the widely adopted multi-channel activation map regression approach. The multi-channel ground-truth activation maps are generated using Gaussian distribution centered on the spatial coordinates of the vertebra centers. The model is trained using L2 loss on the predicted and ground-truth activation maps. 
The produced activation maps are denoted as $G_v \in \mathbb{R}^{W\times H\times L}, \quad v \in \{1,2,\dots,26\}$. Although each activation map channel is trained to activate around the center of the corresponding vertebra, due to the repetitive visual patterns of the vertebrae, it is not uncommon for the heap map to falsely activate on wrong vertebrae, or activate on multiple vertebrae, as shown in Fig. \ref{fig:system}(a).

Standard key point localization methods process the model predicted activation map channels individually (e.g., taking the pixel with the maximum activation or taking the centroid) to obtain the key point detection results. A similar approach has also been adopted to produce vertebra localization and identification results~\cite{chenhao}. Instead of directly processing the activation map channels to obtain vertebra centers, we propose an anatomy-driven processing to achieve robust and accurate vertebra localization and identification, as described in the next two sections.

\subsection{From 3-D to 1-D Spine Rectification}

After obtaining the 3-D vertebra activation maps, we extract the centerline of the spine and aggregate them along the centerline to produce a 1-D vertebra activation signal. The 26 activation maps for individual vertebrae are combined into one activation map:
\begin{equation}
\hat{G} = \sum_{v=1}^{26}{G_v},
\end{equation}
which represents the probability of \textit{any} vertebra center without differentiating their indices. While the individual activation map often falsely activates in wrong vertebrae due to the repetitive image pattern, the activations are typically only around vertebra centers. Therefore, by combining them into one, the centers of all vertebrae are activated, as shown in Fig. \ref{fig:system}(a).

The centerline of the spine is then computed from the combined activation map $\hat{G}$. It is extracted by tracing the mass centers of the axial slices of $\hat{G}$, calculated as the average coordinates of pixels with activation above 0.5. The extracted centerline is denoted as $\mathbf{s}(t) = (x(t), y(t), z(t))$, where $t$ is the arc-length parameterization. Given the spine centerline, the activation maps $G_v$ are warped so that the centerline becomes straight after warping. Specifically, we calculate a moving local coordinate system along the centerline, denoted as $\langle \mathbf{e}_1(t), \mathbf{e}_2(t), \mathbf{e}_3(t) \rangle$. The three axes are chosen as:
\begin{itemize}
	\item $\mathbf{e}_3(t)$: the tangent vector of the curve $\mathbf{s}(t)$.
	\item $\mathbf{e}_2(t)$: the unit vector in the normal plane of $\mathbf{s}(t)$ with the minimum angle to the $y$-axis of the image (i.e., the patient's front direction).
	\item $\mathbf{e}_1(t)$: the cross product of $\mathbf{e}_2(t)$ and $\mathbf{e}_3(t)$.
\end{itemize}
Intuitively, the axes $\mathbf{e}_1(t)$ and $\mathbf{e}_2(t)$ span the normal plane of the spine centerline, where $\mathbf{e}_1(t)$ points at the patient's anterior direction and $\mathbf{e}_2(t)$ directs at the patient's right. 
Given the centerline and the local coordinate systems, we produce spine rectified activation maps $G_v'$ and $\hat{G}'$ by warping $G_v$ and $\hat{G}$, calculated as:
\begin{equation}
G_v'(x,y,z) = G_v(\mathbf{s}(\delta \cdot z) + \mathbf{e}_1(\delta \cdot x) + \mathbf{e}_2(\delta \cdot y)),
\end{equation}
where $G_v(\cdot)$ denotes the linear interpolation of $G_v$ at the given coordinate. This warping operator can be seen as re-sampling $G_v$ in the normal planes of the spine centerline. In the rectified maps, the spine centerline is straight along the $z$ axis, as shown in Fig.~\ref{fig:system}(c). The anterior and right directions of each vertebra are aligned with the $x$ and $y$ axes. 

The rectified activation maps $G_v'$ and $\hat{G}'$ are further processed to produce 1-D signals of vertebra activation, denoted as $Q_v$ and $\hat{Q}$, respectively. Specifically, values in $G_v'$ are summed along the $x$ and $y$ axes, written as
\begin{equation}
Q_v(z) = \sum_{x, y}G'_v(x, y, z).
\end{equation}
Since the activation maps are aggregated in the normal plane of the spine centerline, the produced 1-D signal indicates the likelihood of vertebra centers at given locations $z$ on the spine centerline. The advantages of the 1-D signal are two-fold: 1) by aggregating the activations in the normal plane, the signal of vertebra centers is strengthened, resulting in more distinct activation profile, 2) by reducing the spine localization search space to 1-D, the searching complexity is significantly reduced, making it possible and affordable to adopt more complex optimization approaches. Despite the strengthened activation, false activations in the original activation maps are carried over to the 1-D signal, resulting in false activations in the 1-D signal, as shown in Fig. \ref{fig:system}(c). 

\subsection{Anatomically-constrained Optimization}
\label{ssec:joint_optimization}

\paragraph{Problem Formulation.}

Given the 1-D response signals $\{Q_v\}$ and $\hat{Q}$, we localize and identify the vertebra centers by solving an optimization problem.
Denoting $N$ as the number of detected vertebrae and $v_l$ as the lowest index among them, since the detected vertebrae must be consecutive, their indices can be represented by $[v_l, v_l+N-1]$. The locations of the detected vertebrae are denoted as $\mathbf{k}=\{k_i\}_{i \in [0, N-1]}$, where $i$ is the vertebra's index relative to $v_l$. Therefore, $k_i$ indicates the location of the vertebra with absolute index $v_l + i$. Note that since $N$ can be represented by $\mathbf{k}$, we drop $N$ from the parameters for the sake of notation simplicity. The parameters $(v_l, \mathbf{k})$ are optimized to minimize the following energy function:
\begin{align} 
\mathcal{L}(v_l, & \mathbf{k}) = - \sum_{i=0}^{N-1} {\lambda_{v_l + i} Q_{v_l + i}(k_i)} \nonumber \\ 
& + \sum_{i=2}^{N-2}  R(k_{i} - k_{i-1}, k_{i+1} - k_i). 
\label{eqn:optim}
\end{align}
$Q_{v_l + i}(k_i)$ is the activation value of the vertebra with the absolute index $v=v_l + i$. $R(\cdot,\cdot)$ is a regularization term that encourages the distances between neighboring vertebrae to be similar, written as:
\begin{equation}
R(a, b) = \exp ({\max(\frac{a}{b}, \frac{b}{a})}).
\end{equation}
$\lambda_v$ denotes the weights of the 26 vertebrae. Inspired by the use of anchor vertebrae in~\cite{zhan2012robust}, throughout our experiments, we treat the two vertebrae at the ends of the spine (C1: Cervical-1, C2: Cervical-2, S1: the first Sacrum, S2: the last Sacrum) as anchors and set their weights $\lambda_v$ as 2. For all other vertebrae, the weights are set to 1. Intuitively, these vertebrae (C1, C2, S1, S2) at the ends of the spine have more distinct appearances, and therefore are given more weights than others.

In the above optimization formulation, we jointly search the vertebra centers to maximize the total vertebra activation score while keeping the distances between vertebra centers regulated. The search space of $(v_l, \mathbf{k})$ implicitly imposes a hard constraint that the detected vertebrae must be consecutive with the indices from $v_l$ to $v_l+N-1$.

\paragraph{Optimization Scheme.}
The optimization problem is solved by an initialization step followed by iterative updates. The parameters $(v_l, \mathbf{k})$ are searched in the space: $v_l \in [1, 26 ]$, $k_i \in [0, L]$.
We initialize $v_l=1$ and the vertebra centers $\mathbf{k}$ as the coordinates of local maxima of $\hat{Q}$ sequentially (\ie, $k_{i+1} > k_i$). After the initialization, we iteratively apply three operations to search the parameters, namely 1) \textit{offset}, 2) \textit{fine-tune} and 3) \textit{expansion}.

In the \textit{offset} operation, $v_l$ is optimized via exhaustive search: 
\begin{equation}
v_l \gets \arg \min_{v_l} \mathcal{L}(v_l, \mathbf{k}).
\label{eqn:offset}
\end{equation}

In the \textit{fine-tune} operation, $\{k_v\}$ is optimized via Hill Climbing optimization~\cite{russell2002artificial}:
\begin{equation}
\mathbf{k} \gets \arg \min_{\mathbf{k}} \mathcal{L}(v_l, \mathbf{k}).
\label{eqn:finetune}
\end{equation}
The fine-tune operation adjusts the vertebra centers to minimize the total energy concerning both the individual activation $Q_v$ and the distance regularization.

In the \textit{expansion} operation, a new vertebra center is inserted to $\mathbf{k}$ between $(u, u+1)$. Specifically, the expanded $\mathbf{k}$ is denoted as $E(\mathbf{k}, u)$:
\begin{equation}
    \mathbf{k} \gets E(\mathbf{k}, u) =
    \begin{cases}
    k_i & \text{if $i \leq u$} \\
    (k_i + k_{i+1}) / 2 & \text{if $i = u + 1$} \\ 
    k_{i + 1} & \text{if $i > u$}
    \end{cases}
\label{eqn:expansion}
\end{equation}
The insertion location $u$ is searched by minimizing the energy function below:
\begin{equation}
    u = \arg \min_{u \in [0, N-2]} \mathcal{L}(v_l, E(\mathbf{k}, u)).
\end{equation}
The expansion operation addresses missed vertebrae that are not captured by the local maxima of $\hat{Q}$.

These three operations are iteratively applied until the energy term starts to increase (i.e., indicating convergence). The parameters $(v_l, \mathbf{k})$ associated with the lowest $\mathcal{L}$ during the process are taken as the optimization output. The pseudo code of the proposed optimization scheme is shown in Algorithm \ref{Algo:optimization_new}. After localizing the vertebra centers from the 1-D signals, their coordinates are mapped back to the 3-D CT image following the reverse spatial mapping of the spine rectification to produce the final 3-D localization results. 

\begin{algorithm}[t]
	\SetAlgoLined
	\KwIn{$ Q_{v=1,...,26}(z) $ and $\hat{Q}(z)$}
	$v_l \gets 1$ \;
	$\mathbf{k} \gets$ the coordinates of local maxima of $\hat{Q}(z)$ \;
	$\mathcal{L}_{min} \gets \infty$ \;
	\While{true}{
		$v_l \gets \arg \min\limits_{v_l} \mathcal{L}(v_l, \mathbf{k}) $ \ \ \ // offset \;
		\eIf{$\mathcal{L}(v_{l}, \mathbf{k}) < \mathcal{L}_{min}$}
		{
		    $\mathcal{L}_{min} \gets \mathcal{L}(v_l, \mathbf{k})$ \;
		}
		{
		\textbf{return} $(v_l, \mathbf{k})$ associated with the lowest $\mathcal{L}$ \;
		}
		
		$\mathbf{k} \gets \arg \min\limits_{\mathbf{k}} \mathcal{L}(v_l, \mathbf{k})$ \ \ \ // fine-tune \;
		$u \gets \arg \min\limits_{u \in [0, N-2]} \mathcal{L}(v_l, E(\mathbf{k}, u))$ \;
		$\mathbf{k} \gets E(\mathbf{k}, u)$ \ \ \ // expansion \;
	}
	\KwResult{$(v_l, N, \mathbf{k})$}
	\caption{Optimization}
	\label{Algo:optimization_new}
\end{algorithm}

\begin{table*}[]
	\caption{Comparison of our method with state-of-the-art methods on the SpineWeb test set of 60 CT images. The mean and standard deviation of the localization error (mm) and the identification rate (\%) for different spine regions and their averages are reported.}
	\centering
	    \begin{tabular}{|l|c|c|c|c|c|c|c|c|c|c|c|c|}
		\hline
		\multirow{3}{*}{Method}  &
		\multicolumn{3}{c|}{\textbf{Cervical}} &
		\multicolumn{3}{c|}{\textbf{Thoracic}} &
		\multicolumn{3}{c|}{\textbf{Lumbar}} &
		\multicolumn{3}{c|}{\textbf{All}} \\ \cline{2-13} 
		&
		\begin{tabular}[c]{@{}c@{}}Mean\\ Error\end{tabular} &
		\begin{tabular}[c]{@{}c@{}}Std of\\ Error\end{tabular} &
		\begin{tabular}[c]{@{}c@{}}Id\\ Rate\end{tabular} &
		\begin{tabular}[c]{@{}c@{}}Mean\\ Error\end{tabular} &
		\begin{tabular}[c]{@{}c@{}}Std of\\ Error\end{tabular} &
		\begin{tabular}[c]{@{}c@{}}Id\\ Rate\end{tabular} &
		\begin{tabular}[c]{@{}c@{}}Mean\\ Error\end{tabular} &
		\begin{tabular}[c]{@{}c@{}}Std of\\ Error\end{tabular} &
		\begin{tabular}[c]{@{}c@{}}Id\\ Rate\end{tabular} &
		\begin{tabular}[c]{@{}c@{}}Mean\\ Error\end{tabular} &
		\begin{tabular}[c]{@{}c@{}}Std of\\ Error\end{tabular} &
		\begin{tabular}[c]{@{}c@{}}Id\\ Rate\end{tabular} \\ \hline
		Glocker \etal~\cite{Glocker2013sparseAnnotations} &
		6.81 &
		10.0 &
		88.8 &
		17.6 &
		22.3 &
		61.8 &
		13.1 &
		12.5 &
		79.9 &
		13.2 &
		17.8 &
		74.0 \\ \hline
		McCouat \etal~\cite{McCouat2019VertebraeDA} &
		3.93 &
		5.27 &
		90.6 &
		6.61 &
		7.40 &
		79.8 &
		5.39 &
		8.70 &
		92.0 &
		5.60 &
		7.10 &
		85.8 \\ \hline
		Jakubicek \etal~\cite{jakubicek19learningbased} &
		4.21 &
		 -   &
		 -   &
		5.34 &
		 -   &
		 -   &
		6.64 &
		 -   &
		 -   &
		5.08 &
		3.95 &
		90.9 \\ \hline
		Chen \etal~\cite{chenhao} &
		5.12 &
		8.22 &
		91.8 &
		11.4 &
		16.5 &
		76.4 &
		8.42 &
		8.62 &
		88.1 &
		8.82 &
		13.0 &
		84.2 \\ \hline
		Sekuboy \etal~\cite{btrfly2018} &
		5.90 &
		5.50 &
		89.9 &
		6.80 &
		5.90 &
		86.2 &
		5.80 &
		6.60 &
		91.4 &
		6.20 &
		4.10 &
		88.5 \\ \hline
		Yang \etal~\cite{yangdong2017}&
		5.60 &
		4.00 &
		92.0 &
		9.20 &
		7.90 &
		81.0 &
		11.0 &
		10.8 &
		83.0 &
		8.60 &
		7.80 &
		85.0 \\ \hline
		Liao \etal~\cite{liaohaofu2018} &
		4.48 &
		4.56 &
		95.1 &
		7.78 &
		10.2 &
		84.0 &
		5.61 &
		7.68 &
		92.2 &
		6.47 &
		8.56 &
		88.3 \\ \hline
		Qin \etal~\cite{Qin2020ResidualBM} &
		\textbf{2.20} &
		5.60 &
		90.8 &
		3.40 &
		6.50 &
		86.7 &
		2.90 &
		4.30 &
		89.7 &
		2.90 &
		5.80 &
		89.0 \\ \hline
		Chen \etal~\cite{yizhichen} &
		2.50 &
		3.66 &
		89.5 &
		2.63 &
		3.25 &
		95.3 &
		\textbf{2.19} &
		1.82 &
		\textbf{100} &
		2.56 &
		3.15 &
		94.7 \\ \hline
		Ours &
		2.40 &
		\textbf{1.18} &
		\textbf{96.8} &
		\textbf{2.35} &
		\textbf{1.28} &
		\textbf{97.8} &
		3.19 &
		\textbf{1.69} &
		97.2 &
		\textbf{2.55} &
		\textbf{1.40} &
		\multicolumn{1}{l|}{\textbf{97.4}} \\ \hline
	\end{tabular} \label{tab:comparisons}
\end{table*}

\section{Experiments}
\label{sec:Experiments}

\subsection{Experiment Setup}

\noindent{\bf Dataset.} We have conducted extensive experiments on the public dataset provided by SpineWeb~\cite{csi2014}. The dataset consists of 302 CT scans with vertebra center annotations. This dataset is commonly considered challenging and representative for this task, due to various pathologies and imaging conditions that include severe scoliosis, vertebral fractures, metal implants, and small field-of-view (FOV). In our experiments, we adopt the same dataset split as previous methods~\cite{glockerRegressionForest2012,chenhao,liaohaofu2018,yangdong2017,yizhichen}, where 242 CT scans from 125 patients are used for training and the remaining 60 CT scans are held out for testing.

\noindent{\bf Metrics.} We adopt the two commonly used evaluation metrics: \textit{identification rate} and \textit{localization error}. Identification rate measures the percentage of vertebrae that are successfully identified. A vertebra is considered as correctly identified if the detected vertebra center and the ground truth are mutually the closest and their distance is within 20 mm. Localization error measures the mean and standard deviation of localization errors (in mm) of correctly identified vertebrae. The evaluation metrics are calculated for the vertebrae overall, as well as separately for different spine regions (\ie, cervical, thoracic and lumbar vertebrae).

\subsection{Implementation Details}
We trained our model on a workstation with Intel Xeon CPU E5-2650 v4 CPU @ 2.2 GHz, 132 GB RAM, and 4 NVIDIA TITAN V GPUs. Our method is implemented in PyTorch. The key point localization model is implemented using nnU-Net~\cite{isensee2019automated}~\cite{nnUnetGithub}. CT images are re-sampled to $0.3\times 0.3\times 1.25$ mm spacing. During training, we crop 3-D patches of size $128\times 160\times 64$ voxels from each CT scan as input. For inference, we apply the trained model on non-overlapping patches of the same size to obtain the localization activation maps for the full image. The SGD optimizer with a learning rate of $0.01$, a weight decay of 3e-5 and a mini-batch size of 2 is used to train the model for $1,000$ epochs.

\subsection{Quantitative Comparison with Previous State-of-the-art Methods}
\label{ssec:baseline}

We compare our method with 9 baseline methods, including a classic method with hand-crafted feature~\cite{glockerRegressionForest2012}, multi-stage methods~\cite{McCouat2019VertebraeDA,jakubicek19learningbased}, techniques with data-driven anatomical prior~\cite{liaohaofu2018,yangdong2017,btrfly2018,Qin2020ResidualBM} and methods with anatomy inspired architectures~\cite{chenhao,yizhichen}. The results are summarized in Table~\ref{tab:comparisons}. Overall, our method significantly outperforms all comparative methods, reporting an id. rate of 97.4\% and a mean error of 2.55mm. The closest competitor, Chen \etal~\cite{yizhichen}, reports an id. rate of 94.7\% and a mean error of 2.56 mm. We reduce the id. error rate significantly from 5.3\% to 2.6\%, by absolute 2.7\% (or relative 50.9\%).
When evaluated on three spine regions separately, the id. rates of our method are still better than all comparison methods, except for the lumbar region when compared to Chen \etal~\cite{yizhichen}. On cervical and thoracic spines, our method achieves the highest id. rates of 96.8\% and 97.8\%, respectively.

We note that Chen \etal~\cite{yizhichen} significantly outperforms the other baseline methods in the id. rate. The advantage can be attributed mainly to the adoption of the hard physical constraint imposed by the Markov modeling, which ensures the output to be anatomically plausible. Despite the performance gain, it has a noticeable tendency to achieve higher id. rate on lumbar spine (\ie ranked 1st out of 10) but lower id. rate on cervical spine (\ie ranked 7th out of 10). This is because their method employs Markov model to trace vertebrae from one end of the spine (\ie lumbar) to the other end (\ie cervical). The Markov model successfully regulate the consecutive vertebra indices, which leads to significant performance gain compared to previous methods without such regulation. However, the error can accumulate along with the number of Markov steps as the process goes toward the cervical end. In contrast, our method globally searches and identifies the vertebrae with the constraint of consecutive vertebra indices, which eliminates the directional bias caused by Markov model~\cite{yizhichen} and results in consistent performance in all spine regions.

\subsection{Ablation Study}

\begin{figure*}
    \centering
    \noindent\includegraphics[width=\linewidth]{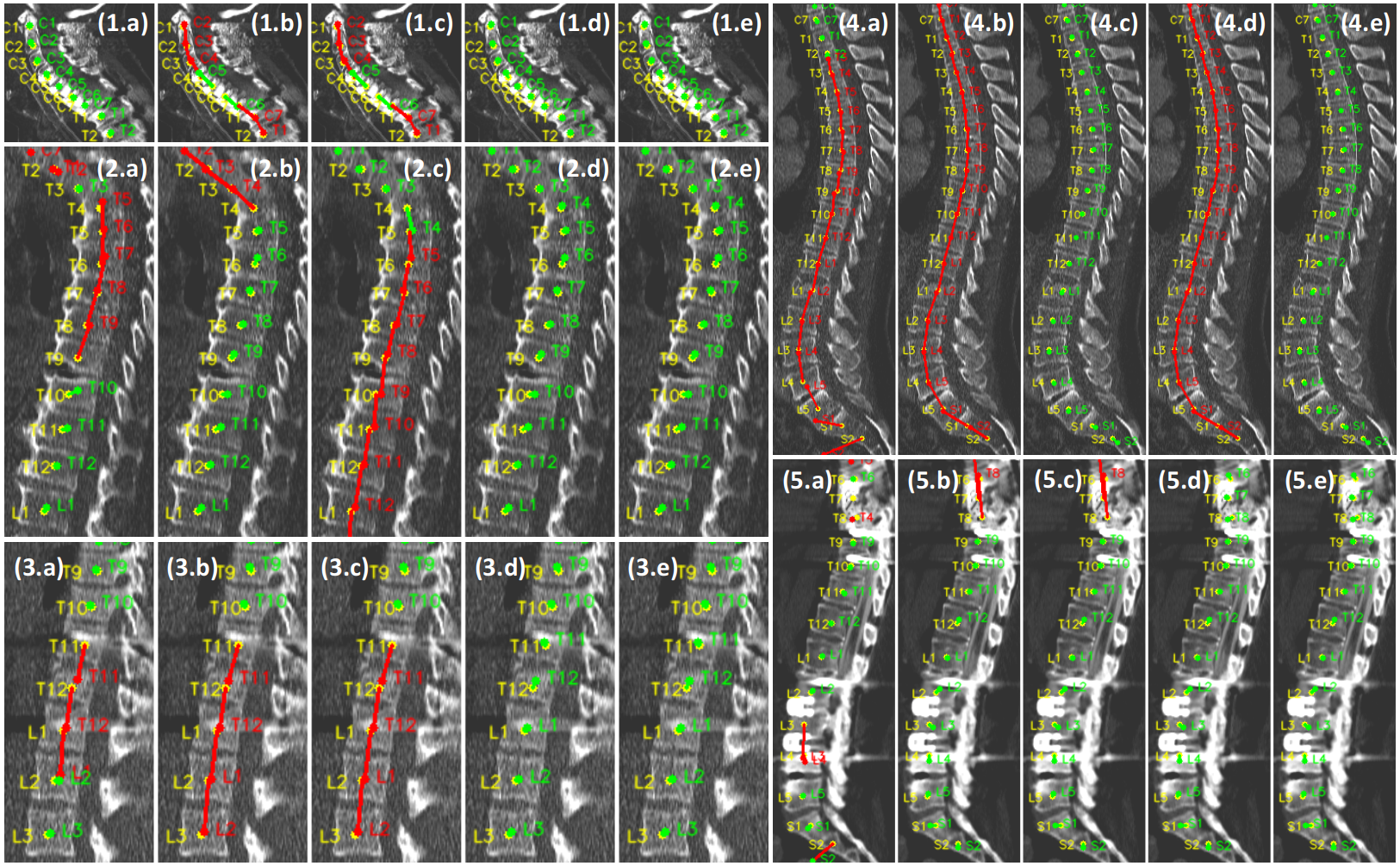} 
    \caption{Visualization of five sets of final results by five methods. \textbf{Dataset (1-5)}: (1) CTs of cervical spine, (2-4) CTs of thoracic and lumbar spine, (5) CTs with metal implant. \textbf{Methods (a-e)}: (a) base, (b) base+rect+order w.o. $\lambda$, (c) base+rect+order, (d) base+rect+optim w.o. $\lambda$, (e) base+rect+optim (ours). The ground-truth vertebra centers are marked by yellow dots and labels. The correct and incorrect predicted vertebra centers are marked in green and red colors, respectively. A line is drawn between the ground-truth and predicted centers of the same vertebra for better visualization of the localization error. }
    \label{fig:example_results}
\end{figure*}

\begin{figure*}[bt!]
	\centering
	\includegraphics[width=\textwidth]{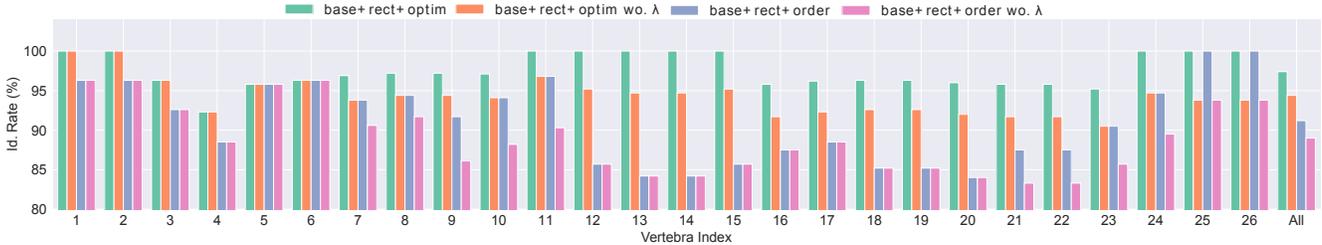}
	\caption{Comparisons of base+rect+order and base+rect+optim with and without using vertebrae weights. The identification rates (\%) for each vertebra and their averages are reported.
	}
	\label{fig:vert_results}
\end{figure*}

\subsubsection{Effects of the Proposed Components}

The spine rectification and anatomically constrained optimization are at the core of our method. In this section, we analyze their effects and behavior via an ablation study of the following alternative methods. The most naive alternative to the iterative optimization is to take the maximum location of individual 3-D activation map $G_v$ as the center of the $v$-th vertebrae, denoted as \textit{base} model. A slightly more sophisticated approach is to take the maximum location of individual 1-D activation signal $Q_v$ as the center of the $v$-th vertebrae. Since this approach employs the spine rectification, it is denoted as \textit{base+rectify}.
In the above two approaches, since there is no constraint applied, physically implausible vertebra orders that violate the anatomy can be produced. A more advanced variation is to take the locations of local maximums of $\hat{Q}$ as candidates of consecutive vertebrae and determine $v_l$ following Equation~\ref{eqn:offset}. This approach ensures the consecutive order of the predicted vertebrae on top of spine rectification, thus referred to as \textit{base+rectify+order}. Note that this approach is equivalent to our method that stops after the {\it offset} operation in the first optimization iteration. Since our method employs both spine rectification and anatomically constrained optimization, it is referred to as \textit{base+rectify+optim}. 

% \usepackage{multirow}
\begin{table*}[]
	\centering
	\caption{Results of the ablation study analyzing the effects of proposed components in our method and the use of vertebra weight $\lambda$. }
	\label{tab:ablation_study}
	    \begin{tabular}{|l|c|c|c|c|c|c|c|c|c|c|c|c|c|}
		\hline
		\multirow{3}{*}{Model} &
		\multicolumn{3}{c|}{\textbf{Cervical}} &
		\multicolumn{3}{c|}{\textbf{Thoracic}} &
		\multicolumn{3}{c|}{\textbf{Lumbar}} &
		\multicolumn{3}{c|}{\textbf{All}} \\ \cline{2-13}
		&
		\multicolumn{1}{c|}{\begin{tabular}[c]{@{}c@{}}Mean\\ Error\end{tabular}} &
		\multicolumn{1}{c|}{Std} &
		\multicolumn{1}{c|}{\begin{tabular}[c]{@{}c@{}}Id\\ Rate\end{tabular}} &
		\multicolumn{1}{c|}{\begin{tabular}[c]{@{}c@{}}Mean\\ Error\end{tabular}} &
		\multicolumn{1}{c|}{Std} &
		\multicolumn{1}{c|}{\begin{tabular}[c]{@{}c@{}}Id\\ Rate\end{tabular}} &
		\multicolumn{1}{c|}{\begin{tabular}[c]{@{}c@{}}Mean\\ Error\end{tabular}} &
		\multicolumn{1}{c|}{Std} &
		\multicolumn{1}{c|}{\begin{tabular}[c]{@{}c@{}}Id\\ Rate\end{tabular}} &
		\multicolumn{1}{c|}{\begin{tabular}[c]{@{}c@{}}Mean\\ Error\end{tabular}} &
		\multicolumn{1}{c|}{Std} &
		\multicolumn{1}{c|}{\begin{tabular}[c]{@{}c@{}}Id\\ Rate\end{tabular}} \\
		\hline
		base &
		2.24 &
		1.25 &
		96.8 &
		2.53 &
		1.53 &
		76.0 &
		2.67 &
		1.66 &
		75.2 &
		2.46 &
		1.48 &
		81.9 \\
% 		\hline
		base+rect &
		2.46 &
		1.64 &
		95.8 &
		2.32 &
		1.63 &
		73.8 &
		3.14 &
		1.70 &
		79.3 &
		2.54 &
		1.68 &
		81.4 \\
% 		\hline
		base+rect+order &
		2.55 &
		1.83 &
		94.2 &
		2.31 &
		1.54 &
		89.4 &
		3.19 &
		1.71 &
		91.0 &
		2.57 &
		1.70 &
		91.2 \\
% 		\hline
		base+rect+optim &
		2.40 &
		1.18 &
		96.8 &
		2.35 &
		1.28 &
		97.8 &
		3.19 &
		1.69 &
		97.2 &
		2.55 &
		1.40 &
		97.4 \\ 
		\hline
		base+rect+order wo. $\lambda$ &
		2.54 &
		1.84 &
		93.7 &
		2.33 &
		1.25 &
		87.2 &
		3.15 &
		1.69 &
		86.9 &
		2.57 &
		1.58 &
		89.0 \\ 
% 		\hline
		base+rect+optim wo. $\lambda$ &
		2.40 &
		1.18 &
		96.3 &
		2.38 &
		1.28 &
		94.1 &
		3.15 &
		1.68 &
		92.4 &
		2.55 &
		1.39 &
		94.4 \\ 
		\hline
	\end{tabular}
\end{table*}

% experiments without enforcing order: FCN base, 1-D signal raw prediction
The results of the ablation study are summarized in Table~\ref{tab:ablation_study} and Fig.~\ref{fig:vert_results}. Visualizations of illustrative image example results are shown in Fig.~\ref{fig:example_results}. The purpose of spine rectification is to enable applying anatomical constraints in the downstream processing. Therefore, employing spine rectification without imposing anatomical constraints does not bring any performance gain, as shown by the comparison between \textit{base} and \textit{base+rect}. By imposing an effective/meaningful constraint of the vertebra order, \textit{base+rect+order} ensures physically plausible results and significantly improves the id. rate over \textit{base+rect} from 81.4\% to 91.2\%. By employing the proposed anatomically constraint optimization, \textit{base+rect+optim} is able to regulate the distance between predicted vertebrae while preserving the physically plausible vertebra order. As a result, the id. rate is further improved from 91.2\% to 97.4\%. We also observe that while the overall id. rate improves significantly, the id. rate for the cervical region is consistently high using the different methods. This is because the cervical vertebrae have a more distinct appearance and can be reliably recognized.

\subsubsection{Effect of the Vertebra Weights $\lambda$}

The vertebra weights $\lambda$ also play an important role by encouraging the optimization to focus more on the vertebrae that can be reliably detected by the key point localization model. To analyze the contribution of the vertebra weights, we conduct an experiment to compare the performances of \textit{base+rect+order} and our method with and without using vertebra weights. As summarized in Table~\ref{tab:ablation_study}, employing vertebra weights leads to improved performance on both \textit{base+rect+order} and our method. In particular, the overall identification rate is improved from 89.0\% and 94.4\% to 91.2\% and 97.4\% on these two methods, respectively. The mean error is not affected much by employing the vertebra weights, which suggests that the vertebra weights have little effect on the accuracy of correctly identified vertebrae.

\subsection{Analysis and Discussion of Failure Cases}
\label{ssec:vis_performance_error}

\begin{figure}[]
	\centering
	\includegraphics[width=\linewidth]{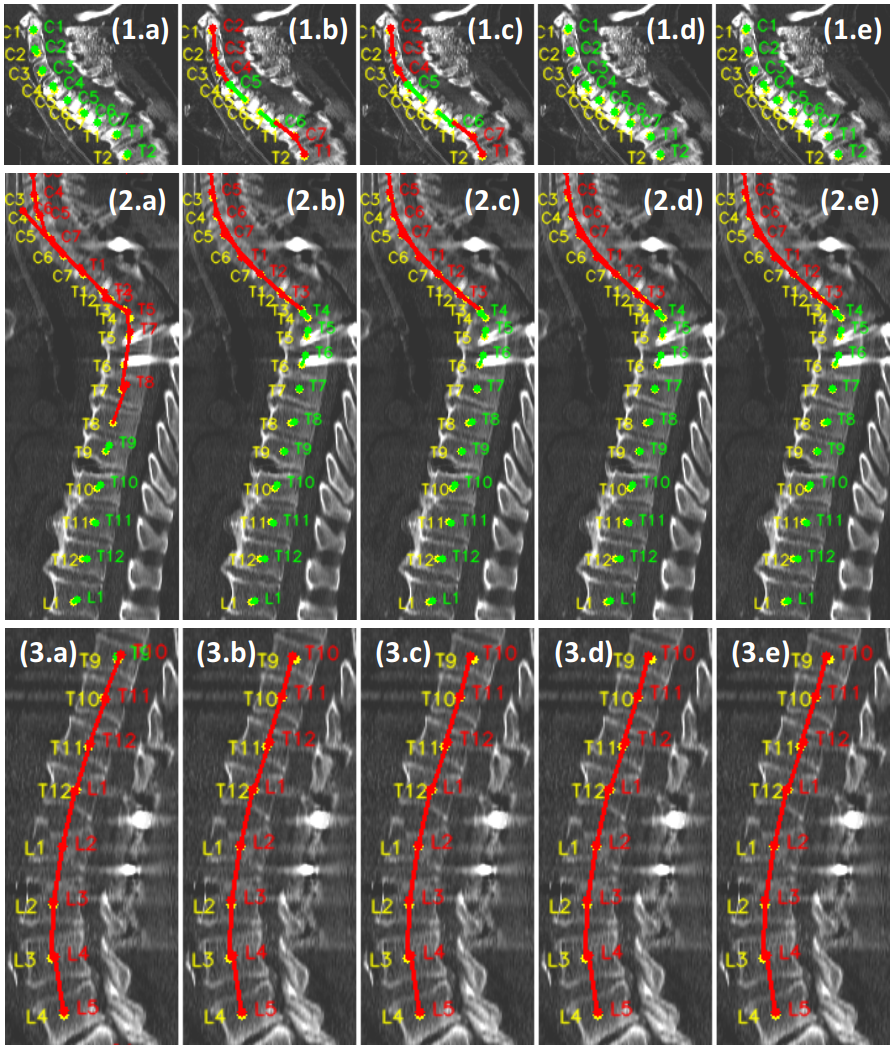}
	\caption{Examples of failure cases. The visualization scheme is the same as in Figure~\ref{fig:example_results}.}
	\label{fig:failure_case}
\end{figure}

In Fig.~\ref{fig:failure_case}, we demonstrate three failure cases of our method. It shows that extreme pathology and/or low quality may degrade the performance of our method. In particular, the first case has severe vertebral compression fractures, which significantly reduces the height of the vertebrae as well as the space margins between them. The second case has low imaging quality, making it difficult to differentiate the boundary between vertebrae. Consequently, we observe missed detection and false positive results in these two cases, respectively. In the last scenario, the vertebra centers are correctly located but labels are off by one. The underlying cause of this failure case is the lack of distinct vertebra that can be reliably recognized. In particular, the more distinct L5 and sacrum vertebrae are not in the field of view. The imaging appearance of T12 vertebrae (the lowest vertebra with rib) is affected by the metal implant.

\section{Conclusion \& Discussion}
\label{sec:Conclusion}

In this paper, we present a highly robust and accurate vertebra localization and identification approach. Based on thorough evaluations on a major public benchmark dataset (i.e., SpineWeb), we demonstrate that by rectifying the spine (via converting and effectively simplifying 3-D detection activation maps into 1-D detection signals) and jointly localizing all vertebrae following the anatomical constraint, our method achieves the new state-of-the-art performance and outperforms previous methods by significantly large quantitative margins. The effectiveness of each proposed algorithmic component has been validated using our ablation studies. 

By analyzing the failure cases, we observe that severe pathologies and extreme imaging conditions may still negatively impact the model's performance on robustness. Therefore, future research efforts should be conducted to further investigate feasible methodologies to improve the robustness against these corner cases of severe vertebral compression fractures, very low imaging contrasts, strong imaging noises such as metal imaging artifacts, and lack of visually distinct anchor vertebrae.

%\vfill
%\pagebreak

%\bibliographystyle{IEEEbib}
%\bibliography{strings,refs}

{\small
\bibliographystyle{ieee_fullname}   %ieee_fullname
\bibliography{strings,refs}
}
\typeout{get arXiv to do 4 passes: Label(s) may have changed. Rerun}
\end{document}